\documentclass[runningheads]{llncs}
\usepackage{graphicx}
\usepackage[utf8]{inputenc} 
\usepackage[T1]{fontenc}
\usepackage{url}
\usepackage{ifthen}
\usepackage{cite}
\usepackage[cmex10]{amsmath} 
\usepackage{amsmath}
\usepackage{booktabs}
\usepackage{graphicx}
\usepackage[utf8]{inputenc}
\usepackage{breqn}
\usepackage{float}
\usepackage{array}
\usepackage{url}
\usepackage{graphicx}
\usepackage{xcolor}
\usepackage{dblfloatfix}
\usepackage{array}
\usepackage{tabto}
\newcolumntype{P}[1]{>{\centering\arraybackslash}p{#1}}
\begin{document}
\title{Cross-domain Variational Capsules for Information Extraction}
%
%
\author{Akash Nagaraj \and
Akhil K \and
Akshay Venkatesh* \and
Srikanth HR}
%

%
\institute{Department of Computer Science, PES University, Bengaluru, India 
\newline
\email{[akashn1897, akhilkred, akshay.venkatesh24]@gmail.com, srikanthhr@pes.edu}}

%
%
\authorrunning{A. Nagaraj et al.}
%

%
\maketitle              
\begin{abstract}
In this paper, we present a characteristic extraction algorithm and the \textit{Multi-domain Image Characteristics Dataset} of characteristic-tagged images to simulate the way a human brain classifies cross-domain information and generates insight. The intent was to identify prominent characteristics in data and use this identification mechanism to auto-generate insight from data in other unseen domains. An information extraction algorithm is proposed which is a combination of Variational Autoencoders (VAEs) and Capsule Networks. Capsule Networks are used to decompose images into their individual features and VAEs are used to explore variations on these decomposed features. Thus, making the model robust in recognizing characteristics from variations of the data. A noteworthy point is that the algorithm uses efficient hierarchical decoding of data which helps in richer output interpretation. Noticing a dearth in the number of datasets that contain visible characteristics in images belonging to various domains, the \textit{Multi-domain Image Characteristics Dataset} was created and made publicly available. It consists of thousands of images across three domains. This dataset was created with the intent of introducing a new benchmark for fine-grained characteristic recognition tasks in the future.

\keywords{Machine reasoning \and Image information \and Capsule networks
\and Variational autoencoders \and Hierarchical decoding.}
\end{abstract}
\section{Introduction}
The machine reasoning domain~\cite{bottou2014machine}, a part of the machine learning umbrella, deals with extracting information from latent data, decoding it and reasoning out the decisions made by machine learning systems. Machine reasoning is a two-step process; the generation of information and the generation of reasoning from this information.

We extract information by training the model on a few domains and testing the model on a new domain. In doing so, the model discovers information from the new domain. Though this might not seem like machine reasoning in the truest sense, it does generate information from latent data. With this paper, we aim to solve a small problem in this vast domain: \textbf{Simulate the way a human brain classifies cross-domain information and generates insight, by identifying prominent characteristics in data and use this identification mechanism to auto-generate insight from data in unseen domains.}

A part of machine reasoning is transfer learning~\cite{pan2009survey}. It stores the knowledge gained from tackling one problem and applies it to another problem which is related to the previous problem solved. Our model incorporates transfer learning to transfer latent information across domains, known as Domain Adaptation~\cite{sun2016return}.

\subsection{Domain Adaptation}
Domain adaptation is a field that deals with machine learning as well as transfer learning. Domain Adaptation can be used when the goal is to learn from one \textit{source} distribution and apply the learning to a different \textit{target} distribution related to the \textit{source}. Scenarios, in which there are multiple source distributions present, are called multi-source domain adaptations. Research being done in this field addresses a major issue - the need to determine a model’s capacity to accurately accept data from a given target domain and label that data accordingly. The challenge arises because the model is trained on a different \textit{source} domain. Unsupervised learning algorithms \cite{barlow} that are implemented without using domain adaptation assume that the examples are independent and identically distributed.

\section{Dataset}
\subsection{Introduction} 
The dataset introduced in this paper, the \textit{Multi-domain Image Characteristic Dataset}~\cite{dataset2020}, consists of thousands of images sourced from the internet. Each image falls under one of three domains - animals, birds or furniture. There are five types under each domain. There are 200 images of each type, summing up the total dataset to 3,000 images. The master file consists of two columns; the image name and the visible characteristics in that image. Every image was manually analysed and the characteristics for each image was generated, ensuring accuracy.

Images falling under the same domain have a similar set of characteristics. For example, pictures under the \textit{Birds} domain will have a common set of characteristics such as the color of the bird, the presence of a beak, wing, eye, legs, etc. Care has been taken to ensure that each image is as unique as possible by including pictures that have different combinations of visible characteristics present. This includes pictures having variations in the capture angle, etc. \newline

\subsection{Why our Dataset is required?} 

At the time of our research, there was a dearth of publicly available datasets that contain visible characteristics in images belonging to various domains. The proposed dataset~\cite{dataset2020} addresses this, as it has the following features:
\begin{itemize}
   \item describes visible characteristics present in every picture.
   \item contains at least hundreds of pictures belonging to multiple domains, and also contains multiple types within each domain. This is crucial to train our model accurately.
   \item contains unique pictures belonging to a type that fall under a certain domain. This is accomplished by collecting pictures that have different combinations of visible characteristics, different angles in which the object was captured, etc.
\end{itemize}

%

\subsection{Training and Testing} 
We recommend a test-train split of 600 samples (20\%) and 2,400 samples (80\%). A \textit{.txt} file with the images to be included in the test and train splits is included, with no overlap between the sets. Following the train-test split as mentioned would help ensure consistency of experiments reported on the \textit{Multi-domain Image Characteristics Dataset}.

\section{Approach}

\subsection{Variational Capsules}
Variational capsules are a combination of capsule networks~\cite{sabour} and variational autoencoders~\cite{doersch2016tutorial}. The capsules generated from capsule networks follow a known prior distribution, and new capsules can be sampled from each of them. They are a natural fit for the model presented in this paper, as they provide a rich representation of image data and are robust to tiny variations in the \textit{decoupled} features of the image.

\subsection{Cross Domain Variational Capsules}
Cross-Domain Variational Capsules are an enhancement to Variational Capsules introduced in the previous subsection. After the latent representation of Variational Capsules is generated for the input image data, this representation is fed to the \textit{Information Decoder}. The \textit{Information Decoder} performs the hierarchical decoding of the rich latent information available from the capsules. In comparison with traditional decoders, our decoder preserves the hierarchical relationship - constructed by the capsules - between features in the data. It leverages the depth of information (in the form of a vector) available for each feature to construct a multi-hot vector identifying the important characteristics from a vocabulary of words spanning all the domains in scope.


The representation can also be leveraged to store cross-domain information to perform information extraction across them. The \textit{Cross-domain Variational Capsule} model is divided into two parts: Creating the latent representation (Variational Autoencoders and Capsule Networks) and Generating insights from that representation (a tailor-made deep network is used for this). A high-level overview of the model can be seen in Fig. 1.
\begin{figure}
\begin{center}
\includegraphics[width = 8cm, height = 4cm]{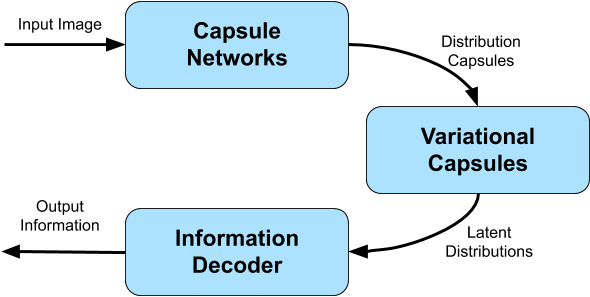}
\caption{Model Design.} \label{ModelDesign}
\end{center}
\end{figure}
\subsection{The Model}

Let $w [lower, higher]$ be a matrix where (lower, higher) are the dimensions of lower-level and higher-level capsules respectively. The depth of the vector (dimensions) is achieved by stacking \textit{m} feature maps together. The vector output of the 32 lower capsules is sent to all the higher-level capsules. 

Essentially, from the squash function, it can be inferred the lower level capsules sends information only to the capsule having the closest centroid to themselves; as it reinforces this connection. It enforces a level of agreement or disagreement between the capsules in different layers. The squash function: 

\begin{center}
    \begin{equation}\label{squash_function}
        v_j = \frac{||s_j^2||}{1 +||s_j^2||}\frac{s_j}{||s_j||}
    \end{equation}
\end{center}

\subsection{Learning Algorithm}

A prediction vector $\hat{u}_{i/j}$ is the prediction from the capsule $i$ to the output of the capsule $j$. If the activity vector $v_j$ is in close \textit{agreement} with the prediction vector $\hat {u}_{i/j}$, we \textit{strengthen} the connection $b_{ij}$. This is the \textit{Routing algorithm} introduced in capsule networks. "Agreement" coefficient:

    \begin{equation}\label{agreement_coeff}
    a_{ij} = < \hat{u}_{i/j}, v_i >
    \end{equation}

The \textit{Routing algorithm} works on inner epochs / iterations which specify the number of times it needs to be run. This is a hyper-parameter to the capsule network model. An epoch starts with $b_{ij} = 0$ for all capsules $i$ in the lower level and corresponding connection capsules $j$ in the higher level.

A normalization function is added to $b_{ij}$. We define 
    \begin{equation}\label{softmax}
    c_{ij} = softmax(b_{ij})
    \end{equation}

An agreement weighted sum is calculated, \begin{equation}\label{agreement_weighted_sum}
    s_{j} = \sum_{i} c_{ij} \hat{u}_{j/i}
\end{equation}

After squishing this sum, we get \begin{equation}\label{after_squishing} v_j = \frac{||s_j^2||}{1 + ||s_j^2||} \hat{s}_{j} \end{equation}

Finally, we update the weight of the connection 
\begin{equation}\label{weighted_connection}
(b_{ij} = b_{ij} + \hat{u}_{j/i} v_j). 
\end{equation}

This process is performed for all pair wise capsule layers.

\subsection{Losses}

The total loss is defined as
\begin{equation}\label{total_loss}
    T_L =  Marginal Loss + \alpha Capsule Loss + \beta KL Divergence Loss.
    \end{equation}
where $\alpha$, $\beta$, and $\gamma$ are constants.

It is important to note that the Reconstruction Loss is not relevant for our model. However for capsule training purposes, we chose to keep it.

\subsubsection{Capsule Loss}

The capsule loss $L_c$ for each capsule is 
    \begin{multline}
        L_c = T_c max(0, m^+ - ||v_c||)^2 + \lambda(1-T_c)max(0,||v_c||-m^-)^2
    \end{multline}

where $T_c$ is $1$ if an object of class $C$ is present (If a \textit{relevant} object is present, the capsule \textit{agrees} with the lower level capsule), $m^+$ is the threshold for $||v_c||$ if $T_c = 1$, $m^-$ is the threshold for $||v_c||$ if $T_c = 0$ and $\lambda$ is a learning hyper-parameter (negative sample loss rate).

\subsubsection{Marginal Loss / Hinge Loss}

The Hinge loss is: 
\begin{equation}\label{Hinge_Loss}
    (L_M = max(0, 1-t.y) 
\end{equation}

where $t$ is the target and $y$ is the output.

\subsubsection{KL Divergence Loss~\cite{hershey2007approximating}}

Let $Z$ be a latent variable, $X$ be a real distribution, $P$ the encoder network, $Q$ the decoder network and $E$ the expectation. 
     \begin{equation}\label{VAO_function}
        log P(X) - D_{KL}(Q(Z|X)||P(Z|X))= E(log P(X|Z)) - D_{KL}(Q(Z|X)||P(Z))
         \end{equation}

(\ref{VAO_function}) is the variational autoencoder objective function. The left-hand side of the objective can be interpreted as lowering the bound of $log P(X)$, which describes our data. The error is the KL Divergence term and lowers the bound of $P(X)$. The maximum likelihood estimate~\cite{myung2003tutorial} (MLE) can be calculated by maximizing $log(P(X|Z))$ and minimizing the difference between the true latent distribution $P(Z)$ and a simple Gaussian distribution $Q(Z|X)$.

Variational autoencoders deal with constructing the underlying distribution of the prior. To achieve this, it uses a reparameterization trick to reconstruct the distribution from the trained $\mu$ and log($\sigma$)$^2$ of the prior. Log variance is used instead of true variance ($\sigma^2$) as it is less volatile and numerically stable.

$D_{KL}$ is to be reduced to $P(Z) = N(0, I)$. Let $Q(Z|X)$ be Gaussian with parameters $\mu(x)$ and $\Sigma(x)$. These are the trainable capsules' mean and log variance. $D_{KL}$ between these two distributions are computed in the close form.
     \begin{equation}\label{D_KL}
     \begin{split}
         D_{KL}[N(\mu(x), \Sigma(x)) || N(0,I)] =  0.5(trace(\Sigma(x)) + \mu(x)^T \mu(x) - k - log det(\Sigma(x)))
         \end{split}
          \end{equation}

where, $k$ is the dimension of the Gaussian distribution, $trace(x)$ is the trace function (sum of diagonal of $X$) and $det(x)$ is the determinant (diagonal of matrix $X$).

    \begin{equation}\label{Final_KL_D_Loss}
        L_{KL} = 0.5 \Sigma_k (\sigma^2(X) + mu^2(X) - 1 - log \sigma^2(X)) 
    \end{equation}

\section{Experiments and Results}

\subsection{Model Evaluation}
\subsubsection{Metrics} 
The model's objective dictates that it is tolerant with noisy characteristics but not with missing ones. Due to this unequal weightage given to false positives and false negatives, \textit{accuracy} is a poor evaluation metric. Hence, the model uses \textit{recall} and \textit{precision} instead. To achieve the objective, the recall must be high, while the precision could be low.

\subsubsection{Evaluation}
To evaluate the performance of the \textit{Cross-domain Variational Capsule} model, we used the \textit{Multi-domain Image Characteristics Dataset}. We have trained the model on 3 domains: Animals, Birds and Furniture. To test the model, we used cross-validation with a 20-80 test-train split.  A simple end-to-end \textit{supervised} training of image vs characteristic gave poor results. We also made sure that capsules were trained sufficiently to accurately generate the rich vector representation for each class. Hence, the model is trained on two levels:
\begin{itemize}
    \item The \textit{Variational Capsule} setup is a typical Capsule Network with output capsules duplicated to be the mean and variances for each capsule. It is trained with the image as the input and the classification as output. This setup uses a modified \textit{Capsule Routing} algorithm to train both sections simultaneously. 
    \item The \textit{Information Decoder} is a hierarchical neural network (where the nodes in a layer are connected to only its parent in the previous layer). It is trained with the image as input and its corresponding characteristics as output. 
\end{itemize}
\subsection{Results}

\begin{table}
\centering
\caption{Model Results}
 \begin{tabular} 
    { |p{3.7cm}|p{3.7cm}|  }
    \hline
    \textbf{Metric} & \textbf{Value}\\
    \hline
    Recall & 0.7666 \\
    Precision &  0.0024\\
    \hline
    \end{tabular}
    \label{table:2}
\end{table}

The results obtained by our algorithm on the \textit{Multi-domain Image Characteristic Dataset} is seen in Table 1. As seen, the value of precision is low, while the value of recall is high because recall depicts the capability of the model to identify relevant characteristics, while precision depicts the proportion of the characteristics the model identifies correctly, to the correct characteristics.

Although the accuracy of the model on a whole is quite low (at about 18\%), considering precision and recall shows that the model can successfully identify characteristics in image data. A point worth noting; F1-score is a metric that finds the balance between precision and recall, and was not a relevant metric to consider in our case, as all the classes had an equal number of data points.

A sample output is seen in Fig.2, showing the probabilities of the characteristics identified in the sample image of a dog from the proposed dataset~\cite{dataset2020}.

\begin{figure}[t]
\begin{center}
\includegraphics[width=14cm, height=6cm]{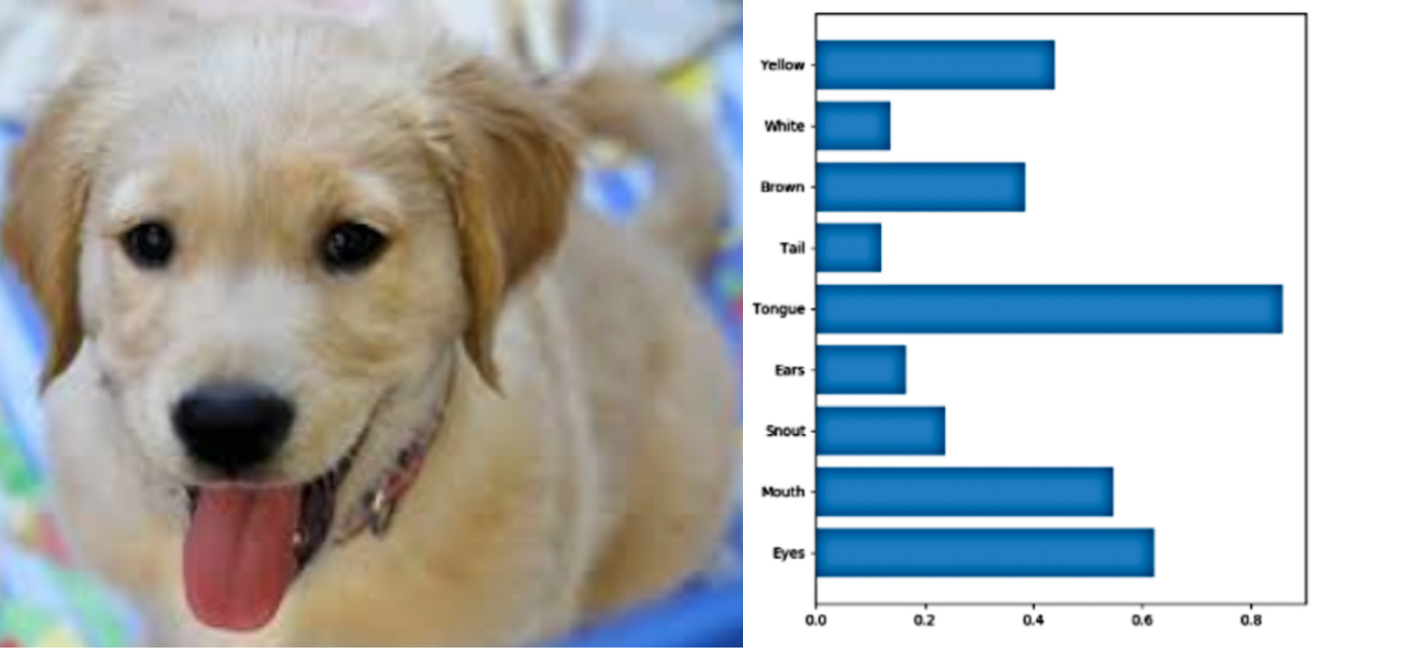}
\caption{Sample Output: Characteristic Identification from a sample image.} \label{CharIdentification}
\end{center}
\end{figure}
\section{Conclusion}

A cross-domain information extraction algorithm using Variational Capsules that learns to extract individual characteristics from image data is proposed. The aim of this algorithm is not to improve an existing model but to satisfactorily solve the relatively recent problem of identifying prominent characteristics of data.

This algorithm preserves the relationship developed between features in capsules, using \textit{hierarchical decoding} as opposed to fully-connected layers. It is also very data efficient, working with a limited number of data points on multi-domain information and is also robust to noise owing to the use of Variational Capsules. Our algorithm was evaluated using the \textit{Multi-Domain Image Characteristics Dataset}, confirming that it successfully extracts characteristics (or information in general) from image data. The algorithm can also work on any form of data supported by capsules. Potential applications of our algorithm are numerous as information extraction is used in a wide number of fields. Image characteristic extraction is also very versatile and is used in a plethora of fields ranging from autonomous driving to astronomy. 

\subsection{Future Enhancements}

Future enhancements include experimentation with different data formats (audio, text, etc.) and \textit{characteristic recognition} methods. Applying the above algorithm to different data formats, and extracting characteristics from the data, we aim to best represent the underlying characteristics of all formats of data. An additional improvement would be to improve the efficiency and speed of the proposed algorithm, drawing inspiration from similar real-time approaches~\cite{nagaraj2018real}.

\bibliographystyle{splncs04}
\bibliography{bibli1}

\end{document}